\documentclass{article}

\usepackage{PRIMEarxiv}

\usepackage[utf8]{inputenc} 
\usepackage[T1]{fontenc}    
\usepackage{hyperref}       
\usepackage{url}            
\usepackage{booktabs}       
\usepackage{amsfonts}       
\usepackage{nicefrac}       
\usepackage{microtype}      
\usepackage{lipsum}
\usepackage{fancyhdr}       
\usepackage{graphicx}       
\graphicspath{{media/}}     

\usepackage{graphicx}
\usepackage{color}

\usepackage{amsmath, amssymb, bm}
\usepackage{soul, xcolor}
\usepackage{amsmath}
\usepackage{amsfonts}
\usepackage{amssymb}
\usepackage{fancyhdr}
\usepackage{subcaption}
\usepackage{geometry}
\usepackage[english]{babel}
\usepackage{csquotes}
\usepackage{xcolor} 
\usepackage{listings}
\usepackage{soul}
\usepackage{float}
\usepackage[linesnumbered,ruled,vlined]{algorithm2e}
\usepackage{multirow}
\usepackage{verbatim}
  
\usepackage{subcaption}
\usepackage{paralist}
\usepackage{enumitem}

\newlist{notes}{enumerate}{1}
\setlist[notes]{label=\textbf{Note:}, leftmargin=*}

\DeclareMathOperator{\EX}{\mathbb{E}} 
\DeclareMathOperator*{\argmin}{\arg\!\min} 

\title{Denoising Diffusion Planner: Learning Complex Paths from Low-Quality Demonstrations}

\author{
  Michiel Nikken$^1$, Nicolò Botteghi$^2$, Wesley Roozing$^1$, Federico Califano$^1$ \\
  $^1$ Robotics and Mechatronics (RaM), University of Twente, Enschede \\
  $^2$ Mathematics of Imaging and AI (MIA), University of Twente, Enschede \\
  \texttt{Contacts: \{n.botteghi,f.califano\}@utwente.nl} \\
}

\begin{document}
\maketitle

\begin{abstract}
 Denoising Diffusion Probabilistic Models (DDPMs) are powerful generative deep learning models that have been very successful at image generation, and, very recently, in path planning and control. In this paper, we investigate how to leverage the generalization and conditional sampling capabilities of DDPMs to generate complex paths for a robotic end effector. We show that training a DDPM with synthetic and low-quality demonstrations is sufficient for generating nontrivial paths reaching arbitrary targets and avoiding obstacles. Additionally,  we investigate different strategies for conditional sampling combining classifier-free and classifier-guided approaches. Eventually, we deploy the DDPM in a receding-horizon control scheme to enhance its planning capabilities. The Denoising Diffusion Planner is experimentally validated through various experiments on a Franka Emika Panda robot.  Videos of our experiments can be found \href{https://1drv.ms/f/c/4b7f21802058fb05/Eq4SFEAvfIlIhsqF6RoGT1ABgGEJd0UXHPpkpZlwmFt2mA?e=AVZqCj}{here}.
\end{abstract}

\twocolumn
\section{Introduction}

Path planning is a fundamental aspect of almost any robotic task. It consists in finding a collision-free path between a starting state and a goal state, such that the system can be driven to the desired target by tracking this path using a low-level controller. Designing path planning algorithms is a difficult problem, because robotic environments are typically high-dimensional, complex, and dynamic.

Traditionally, to address the task of path planning, combinatorial algorithms have been developed. These algorithms are complete, meaning that they will find a solution in a finite amount of time, if a solution exists \cite{Schwartz1988AAlgorithms}. However, for complex environments it is often preferable to trade completeness for efficiency. To this end, sampling-based methods \cite{Karaman2011Sampling-basedPlanning} have been successfully developed. Examples of traditional sampling-based methods are the probabilistic roadmap method (PRM) \cite{Kavraki1996ProbabilisticSpaces} and rapidly exploring random trees (RRTs) \cite{Lavalle1998Rapidly-exploringPlanning}.

More recently, advances in reinforcement learning (RL) \cite{Sutton2018ReinforcementIntroduction} have introduced
a different approach to planning. RL seeks to find the optimal control strategy that maximizes the expected cumulative future rewards by interacting with an unknown environment. Utilizing deep neural networks as function approximators to solve RL problems is known as deep reinforcement learning (DRL) \cite{Sutton2018ReinforcementIntroduction}, which has allowed RL to scale to high-dimensional problems \cite{Mnih2013PlayingLearning, Mnih2015Human-levelLearning}. Iteratively executing the actions generated by the policy while updating the current state estimates using observations from the environment makes for a reactive planner \cite{Konar2013ARobot,Low2019SolvingQ-learning, lei2018dynamic, yao2020path, botteghi2020reward}.

Despite their successful application in different path-planning problems, the previously-mentioned traditional sampling-based methods and DRL methods have several drawbacks. For example, it is not obvious how to plan a complex behavior that meets conditions other than obstacle avoidance using traditional sampling-based planners. On the other hand, DRL methods learn a richer model that may include system dynamics and arbitrary reward functions. However, applying these methods for planning requires autoregressively using their model's one-step predictions, which allows model imperfections to compound over time \cite{Asadi2019CombatingModel, Xiao2019LearningLearning}.

Instead, the problem of path planning can be addressed through the paradigm of planning as inference (PAI) \cite{Attias2003PlanningInference, Botvinick2012PlanningInference}. Differently from explicit graph search or autoregressive prediction, the future is modeled as a joint probability distribution over states, actions and rewards. Sampling from this distribution yields a sequence of states and actions that represent a possible path. Retrieving just any possible path is not very useful by itself, but if this path is known to meet certain conditions, like obstacle avoidance or achieving certain rewards, this sample can be used as a plan. In PAI, such plan can be sampled from a conditional probability distribution of possible futures. This process of conditional sampling is key to generate a plan that exhibits desirable behavior.

A particularly recent class of models that fit the PAI view is the Denoising Diffusion Probabilistic Model (DDPM) \cite{Sohl-Dickstein2015DeepThermodynamics, Ho2020DenoisingModels}. DDPMs are generative probabilistic models that have been remarkably successful in image generation \cite{Dhariwal2021DiffusionSynthesis}. They are trained by iteratively adding random noise to the training data and learning the reverse process that iteratively denoises the data to retrieve the original sample. In this way, one can sample from the data distribution carrying the characteristics of the training data by first sampling noise from a Gaussian distribution and then applying the denoising process.

\begin{figure*}
    \centering
    \includegraphics[width=\linewidth]{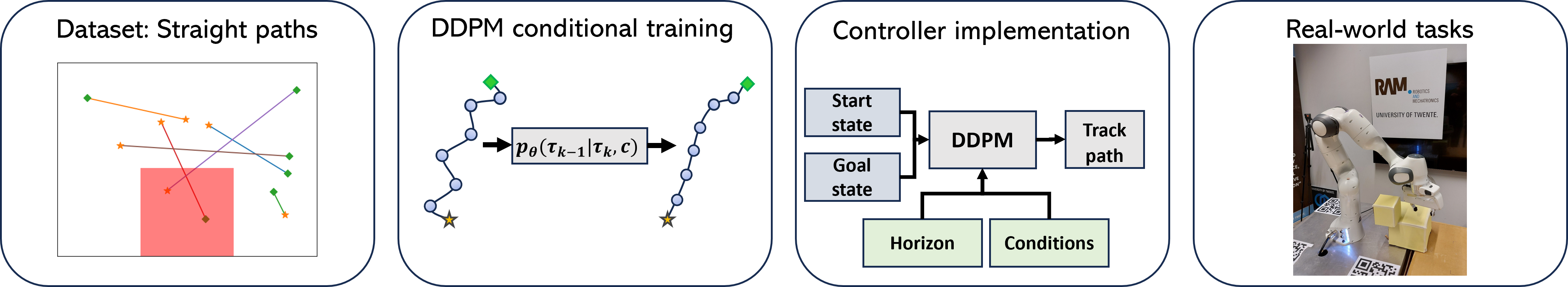}
    \caption{Overview of the DDP pipeline from low-quality and synthetic demonstration to real-world tasks.}
    \label{fig:overview}
\end{figure*}

The generative capabilities of DDPMs can also be used to generate plans. Janner et al. \cite{Janner2022PlanningSynthesis} trained DDPMs on an offline collection of state-action sequences, with associated rewards. In \cite{Janner2022PlanningSynthesis} it was shown that DDPMs exhibit good generalization and long-horizon planning abilities, even for sparse rewards. Furthermore, they showed that DDPMs can generate novel trajectories by locally stitching together sequences from the training distribution. Moreover, Ajay et al. \cite{Ajay2023IsDecision-Making} found that DDPMs can generate plans that meet multiple conditions even though these conditions were never met simultaneously in the training data. These properties of DDPMs make them potentially powerful and versatile planners. Real-world demonstrations of using DDPMs for robot control have shown promising results \cite{Chi2023DiffusionDiffusion, Liu2024DiPPeR:Robots, Carvalho2023MotionModels}, which inspires this work to further investigate the novel aspects of using DDPMs to solve robotic planning problems. Carvalho et al. \cite{Carvalho2023MotionModels} propose a motion planner in the joint space using DDPMs. They use expert examples produced by an optimal motion planning algorithm to train a DDPM. During inference, they use the gradients of a cost function to bias the samples towards regions of low cost at every denoising step. Saha et al. \cite{Saha2024EDMP:Planning} expand on this work by using an ensemble of cost functions to create a planner that can generate collision-free trajectories in a variety of environments. Chi et al. \cite{Chi2023DiffusionDiffusion} use DDPMs to directly learn a policy that takes visual observations as its input and produces an action sequence as an output. In their work, they introduce closed-loop planning with DDPMs by implementing a receding-horizon control scheme.
DDPM can also be used for safety-critical planning in combination with control barrier functions \cite{ames2019control} to generate optimal and safe plans \cite{xiao2023safediffuser, Botteghi2023TrajectoryModels}.

In this work, we investigate how to use a DDPM to create a path planner for an end effector of a robotic arm (see Figure \ref{fig:overview}). 
In particular, to exploit the DDPM's generalization capabilities, we train the DDPM using only synthetically-generated low-quality demonstrations, namely straight paths with associated returns. This is a key difference from \cite{Carvalho2023MotionModels, Saha2024EDMP:Planning, xiao2023safediffuser, Botteghi2023TrajectoryModels, gerges2024invisible}, which use expert demonstrations.  Inspired by \cite{Chi2023DiffusionDiffusion}, we include closed-loop planning to enhance the planning capabilities of the DDPMs. Eventually, we investigate and empirically evaluate different strategies for conditional sampling and generation of optimal paths. We name our approach Denoising Diffusion Planner (DDP).
The main contributions of this work are as follows:
\begin{enumerate}
    \item A DDPM-based path planner for a robotic end-effector, not relying on a dynamic model of the robot and using only synthetic and low quality demonstrations;
    \item Experimental validation on a 7-DoF robot manipulator, comparing five different conditional-sampling strategies.
\end{enumerate}

\section{Background}\label{sec:background}

\subsection{Denoising Diffusion Probabilistic Models}\label{sec:ddpm}
Denoising Diffusion Probabilistic Models (DDPM) \cite{Sohl-Dickstein2015DeepThermodynamics,Ho2020DenoisingModels} are a class of generative deep learning models that can learn data distributions according to a given training dataset in a way that allows for sampling from the learned distribution. DDPMs are characterized by a forward process and a reverse process. In the forward process, Gaussian noise is added to the training data iteratively. The forward process is a discrete Markov chain, i.e., a memoryless process in which the previous state of the chain $\bm{x}^{k-1}$ is sufficient to determine the current state $\bm{x}^k$. In general, given a noise-free sample $\bm{x}^0$ from dataset $\mathcal{D}$, the forward process can be expressed as follows:
\begin{equation}\label{eq:diffusion_steps}
    q \left(\bm{x}^{K} \middle| \bm{x}^0\right) = \prod^K_{k=1} q \left(\bm{x}^k \middle| \bm{x}^{k-1}\right)\, ,
\end{equation}
where $k=0, 1, ... , K$ is the index of the diffusion step and $q \left(\bm{x}^k \middle| \bm{x}^{k-1}\right)$ the Markov diffusion kernel. This kernel is a probability distribution describing the probability of $\bm{x}^k$ for a given $\bm{x}^{k-1}$. The kernel is chosen such that its repeated application will transform the initial data distribution $q\left(\bm{x}^{0}\right)$ into a Gaussian distribution $\mathcal{N}\left(\bm{0}, \bm{I}\right)$:
\begin{equation}\label{eq:euc_diffusion_step}
    q \left(\bm{x}^k \middle| \bm{x}^{k-1}\right) = \mathcal{N}\left(\sqrt{1 - \beta^k}\bm{x}^{k-1}, \beta^k \bm{I}\right)\, ,
\end{equation}
 where $\beta^k \in \left(0, 1\right)$ is the variance in the forward process. Assuming that $q\left(\bm{\tau}^0\right)$ has been normalized to unit variance, scaling the mean of the Gaussian kernel by $\sqrt{1-\beta^k}$ will keep the variance of $q\left(\bm{\tau}^k|\bm{\tau}^0\right)$ unity. Hence, $q\left(\bm{\tau}^{K}\right)$ will resemble a standard normal distribution for sufficiently large number of diffusion steps $K$.

The reverse process is, again, a Markov chain, parameterized by a parameter vector $\bm{\theta}$, that can be written in as:
\begin{equation}\label{eq:reverse}
\begin{split}
    p_{\bm{\theta}} \left(\bm{x}^{0}\right) & = \prod^K_{k=1} p{_{\bm{\theta}}} \left(\bm{x}^{k-1} \middle| \bm{x}^k\right), \\
    p{_{\bm{\theta}}} \left(\bm{x}^{k-1} \middle| \bm{x}^k\right) & = \mathcal{N}\left(\bm{\mu_\theta} \left(\bm{x}^k, k\right), \bm{\Sigma_\theta}\left(\bm{x}^k, k\right)\right)\, ,
\end{split}
\end{equation}
where $\bm{\mu_\theta} \left(\bm{x}^k, k\right)$ is learned from data, using a deep neural network with parameters indicated by $\boldsymbol{\theta}$, and the covariance is fixed to a cosine schedule, i.e., $ \bm{\Sigma}_\theta\left(\bm{x}^k, k\right) =  \sigma^k \bm{I}$, as proposed by Nichol and Dhariwal \cite{Nichol2021ImprovedModels}.

Ho et al. \cite{Ho2020DenoisingModels} found that the Gaussian noise kernel allows for a closed-form description of the forward process:
\begin{equation}\label{eq:forward_closed_form}
    q \left(\bm{x}^k \middle| \bm{x}^0 \right) = \mathcal{N}\left(\sqrt{\Bar{\alpha}^k}\bm{x}^0, \left(1 - \Bar{\alpha}^k\right)\bm{I}\right),
\end{equation}
where $\Bar{\alpha}^k=\prod^k_{s=1}1-\beta^s$. The reverse process can also be written in closed-form, expressing the probability of $\bm{x}^{k-1}$ when $\bm{x}^{k}$ and $\bm{x}^{0}$ are given:
\begin{equation}\label{eq:euc_posterior}
    \begin{split}
        &p\left(\bm{x}^{k-1}|\bm{x}^k, \bm{x}^0\right) = \mathcal{N}\left(\Tilde{\bm{\mu}}^k\left(\bm{x}^k, \bm{x}^0\right), \Tilde{\beta}^k \bm{I} \right)\, , \\
        &\Tilde{\bm{\mu}}^k\left(\bm{x}^k, \bm{x}^0\right) =  \frac{\sqrt{\Bar{\alpha}^{k-1}}\beta^k}{1 - \Bar{\alpha}^k} \bm{x}^0 + \frac{\sqrt{\Bar{\alpha}^{k}}\left(1 - \Bar{\alpha}^{k-1}\right)}{1 - \Bar{\alpha}^k} \bm{x}^k\, , \\
        &\Tilde{\beta}^k = \frac{1 - \Bar{\alpha}^{k-1}}{1 - \Bar{\alpha}^k}\beta^k\, ,
    \end{split}
\end{equation}
Equation \eqref{eq:forward_closed_form} can be rewritten as a function of the added noise $\boldsymbol{\epsilon}$ to $\bm{x}_k\left(\bm{x}_0, \bm{\epsilon}\right) = \sqrt{\Bar{\alpha}_k}\bm{x}_0 + \sqrt{1 - \Bar{\alpha}_k} \bm{\epsilon}$, where $\bm{\epsilon} \sim \mathcal{N}\left(\bm{0}, \bm{I}\right)$. Using this reparameterization, Ho et al. \cite{Ho2020DenoisingModels} derived a simplified training objective for a neural network $\bm{\epsilon}_{\bm{\theta}}$ that estimates the noise $\bm{\epsilon}$:
\begin{equation}\label{eq:training_objective}
    L\left(\boldsymbol{\theta}\right) = \EX_{k,\bm{\epsilon},\bm{x}^0}\left[\lVert \bm{\epsilon} - \bm{\epsilon}_\theta\left(\bm{x}^k, k\right) \rVert^2 \right].
\end{equation}

\subsubsection{Conditional Sampling with DDPMs}
In general, the reverse process is trained to turn samples from a standard normal distribution into samples that resemble those in the training dataset. Oftentimes, we have additional demands to the samples that are generated, i.e., the samples need to meet certain conditions. This is achieved through conditional sampling. A condition vector $\bm{c} \in \mathbb{R}^n$ encodes a property of the samples, for example, a class. Different techniques have been proposed for condition sampling, such as classifier guidance \cite{Dhariwal2021DiffusionSynthesis} and classifier-free guidance \cite{Ho2022Classifier-FreeGuidance}. In this work, we opt to use classifier-free guidance rather than classifier guidance, since the latter involves training an additional classifier model, which increases the training cost and complexity.

Classifier-free guidance guides the reverse process to produce samples that meet the condition using an additional input to the DDPM. The condition $\bm{c}$ is added to $ \bm{\epsilon}_\theta\left(\bm{x}^k, k\right)$ in Equation \eqref{eq:training_objective} and the noise prediction is subsequently rewritten as:
\begin{equation}\label{eq:classifer_free_noise}
\begin{split}\bm{\Tilde{\bm{\epsilon}}_\theta}\left(\bm{x}^k, k, \bm{c}\right) = &\bm{\epsilon}_\theta\left(\bm{x}^k, k, \varnothing \right) + \\ &w \left(\bm{\epsilon}_\theta\left(\bm{x}^k, k, \bm{c}\right) - \bm{\epsilon}_\theta\left(\bm{x}^k, k, \varnothing \right)\right)\, ,    
\end{split}
\end{equation}
where $w$ is a weight factor to balance unconditional and conditional generation and $\varnothing$ is a null-token used as the unconditional embedding. To train the neural network, a random binary mask is applied to the condition embedding that is passed to the network at each training step. In this way, the network cannot use the conditional information for every prediction. Therefore, the DDPM simultaneously learns the conditional and unconditional relation.

Additionally, DDPMs are capable of inpainting. Fixing a component of $\bm{x}^k$ to a desired value at every diffusion step $k=K, K-1, ... , 0$ causes the DDPM to inpaint about the desired component.

\section{Denoising Diffusion Planner}
In this section, we introduce our Denoising Diffusion Planner (DDP) approach based on DDPM, conditional sampling, and closed-loop control.

In this work, each path is discretized into a sequence of poses equally spaced in time, although the methodology applies to any parametrisations. The term ``trajectory" is often used in literature for time-parameterized paths, and therefore we will use ``path" and ``trajectory" interchangeably from now on. 

We indicate a path with $\bm{\tau} = \left(H_0 , H_1, ..., H_T\right)$, where $H$ is an element of the Lie-group SE(3) \cite{Sola2018ARobotics} which encodes the position and orientation of a frame with respect to a chosen reference frame.
In particular, a pose $H_t$ can be described by a matrix that is composed of a rotation matrix $R_t$, encoding the orientation, and a vector $\bm{p}_t$, encoding the position \cite{Sola2018ARobotics}:
\begin{equation}\label{eq:Hmat}
    H_t = 
    \begin{bmatrix}
        R_t & \bm{p}_t\\
        \bm{0}^{\top} & 1
    \end{bmatrix}\, ,
\end{equation}
where $\top$ refers to the transpose operator and $\bm{0}$ is a column vector of three zeros.

\begin{notes}
    \item There are two indices to keep track of: the time index and the diffusion index. We use the convention of Janner et al. \cite{Janner2022PlanningSynthesis} to use subscripts to denote the time index and superscripts to denote the index of the diffusion process. For example, a pose at time step $t$ and noise step $k$ is hence written as $H^k_t$.
\end{notes} 

\begin{figure}[h!]
    \centering
    \includegraphics[width=1.0\linewidth]{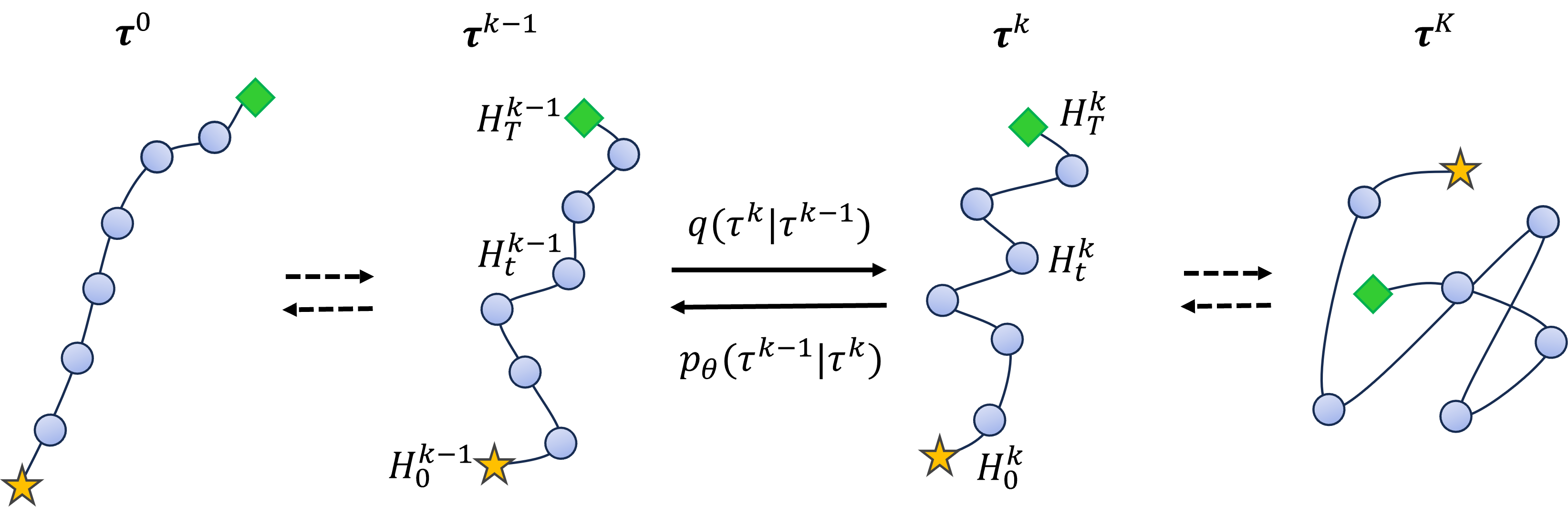}
    \caption{Forward process and reverse diffusion process for a path $\boldsymbol{\tau}$. }
    \label{fig:diffdenoise}
\end{figure}

To enable the DDP to generate paths, the diffusion process is defined for the state trajectory $\bm{\tau}$, giving $q \left(\bm{\tau}^{K} \middle| \bm{\tau}^0\right)$ for the forward process and $p_\mathbf{\theta} \left(\bm{\tau}^{0}\right)$ for the reverse process. Figure \ref{fig:diffdenoise} depicts the diffusion process for a path. This is achieved by choosing coordinates for $H_t$ and stacking all entries of $\bm{\tau}$ along the time dimension. The path $\bm{\tau}$ is then represented by a 2D matrix to which all equations in Section \ref{sec:ddpm} apply by setting $\boldsymbol{\tau}^k = \bm{x}^k = \left[ H^k_0, H^k_1, ..., H^k_T \right]$.

\subsection{Path Planning through Conditional Sampling}
\begin{figure}[h!]
    \centering
\includegraphics[width=0.8\linewidth]{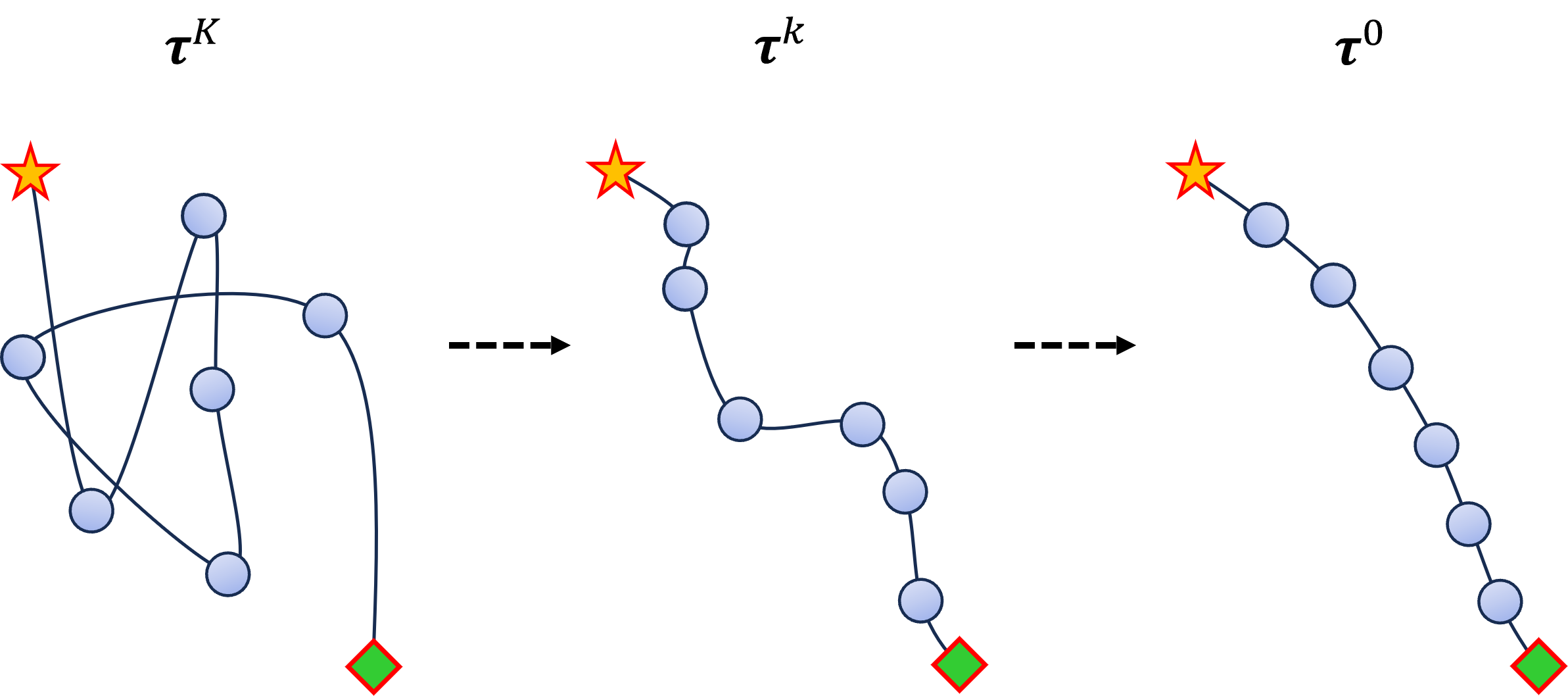}
    \caption{Inpainting is used to set the start and goal poses of a generated path by fixing the start and goal poses (green square and yellow star, respectively) for all $k=K,K-1,...,0$.}
    \label{fig:inpainting}
\end{figure}

Sampling Gaussian noise and applying the reverse process yields entire paths that match the distribution of the data that the DDPM is trained on, without requiring autoregression along the time dimension. To ensure that the generated paths go from the starting pose to the goal pose, inpainting is used. That is, $H^k_0$ and $H^k_T$ are fixed for all diffusion steps $k$, as is displayed in Figure \ref{fig:inpainting}.
To sample paths that avoid obstacles, two conditional sampling techniques are investigated. Firstly, we use classifier-free guidance. We use as condition the return of a path given by the sum of discounted rewards:
\begin{equation}
    c = \sum^T_{t=0} \gamma^t r_t(H_t)\, ,
    \label{eq:returns}
\end{equation}
where $\gamma$ is the discount factor and $r_t(H_t)$ a reward based on the distance to the obstacles, see \eqref{eq:r_dense} and \eqref{eq:r_sparse}. The condition vector $\bm{c}$, in \eqref{eq:classifer_free_noise}, is obtained by encoding the return $c$ using a small neural network $\phi_{\boldsymbol{\theta}}$ that is trained with the DDPM such that $\bm{c}=\phi_{\boldsymbol{\theta}}(c)$. Conversely, during unconditional training and generation the embedding vector is set to zero, i.e., $\bm{c}=\bm{0}$. Using classifier-free guidance we can sample paths with high returns, i.e., steering away from obstacles.

In related work, Carvalho et al. \cite{Carvalho2023MotionModels} introduced a different way of conditional sampling, inspired by classifier-guided sampling \cite{Dhariwal2021DiffusionSynthesis}. They define a cost function $J\left(\bm{\tau}^k\right)$,  containing different terms to penalize collisions and promote smooth trajectories towards the target. During inference, they optimize this cost function  by adjusting the mean of the reverse process\footnote{This approach is analogous to the classifier guidance but without the need for training an additional DDPM classifier.} in \eqref{eq:reverse} using the gradients of the cost function as follows:
\begin{equation}\bm{\Tilde{\mu}_\theta}\left(\bm{\tau}^k,k\right) = \bm{{\mu}_\theta}\left(\bm{\tau}^k,k\right) - \nabla_{\bm{\tau}^k}J(\bm{\tau}^k)|_{\bm{\tau}^k= \bm{{\mu}_\theta}\left(\bm{\tau}^k,k\right)}.
\end{equation}
This method will hereinafter be referred to as "cost-guided sampling", or simply "cost guidance". We will also investigate this method of conditional sampling to achieve obstacle avoidance, as well as the combined application of classifier-free guidance and cost guidance.

Furthermore, the probabilistic nature of DDPMs allows for a Monte Carlo approach to further optimize the paths. Instead of sampling a single path, DDPMs can be used to sample $n$ different paths using the same conditions and selecting the path in the batch that minimizes a cost function $C\left(\bm{\tau}^k\right)$:
\begin{equation}
    \bm{\tau}^* = \argmin_{\bm{\tau}} (C(\bm{\tau})).
\end{equation}

\subsection{Closed-loop Planning}\label{sec:closed-loop}
\begin{figure}[h!]
    \centering
    \includegraphics[width=1.0\linewidth]{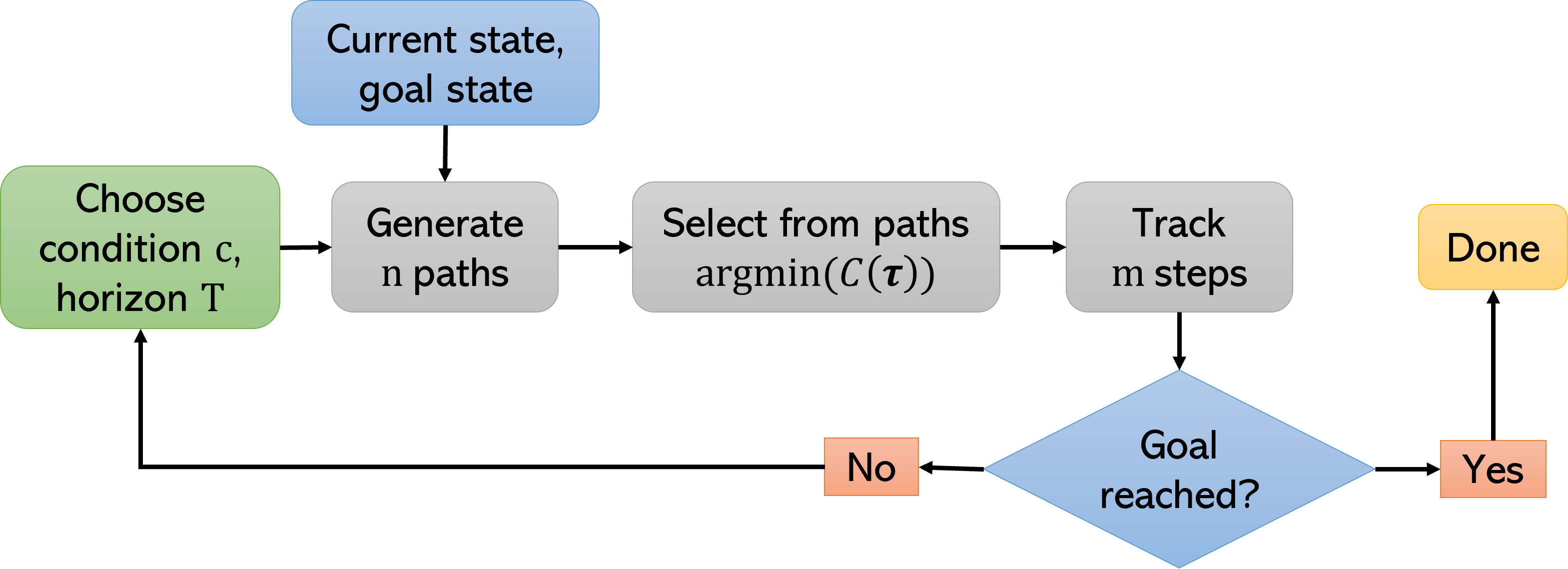}
    \caption{Block diagram of a closed-loop control configuration for a DDPM planner.}
    \label{fig:closedloop}
\end{figure}

Although DDPMs do not have an inherent sense of temporal ordening, the quality of a path may still degrade along the time dimension due to discounting of the rewards, in the sense that future poses carry less weight in the reward structure (see \eqref{eq:returns}). This issue can be addressed with a closed-loop approach to path planning. Rather than sampling a path and tracking it entirely, one can sample a path, track it for $m$ steps, and then resample to update the rest of the path. Upon resampling, the planning horizon is shifted to achieve receding-horizon control, like Chi et al. \cite{Chi2023DiffusionDiffusion} did. A block diagram of the process is shown in Figure \ref{fig:closedloop}. 
This approach resembles Model Predictive Control \cite{garcia1989model} in the sense that we have a model that predicts future states and optimizes some cost function over a finite horizon.

\section{Experimental Design}
The following section describes various simulations and experiments performed with the DDP. Training and inference settings that are fixed throughout this work are in Table \ref{tab:settings}.
\begin{table}[h!]
\centering
\caption{Hyperparameters used in our experiments.}
\label{tab:settings}
\begin{tabular}{@{}ll@{}}
\toprule
Setting & Value \\ \midrule
Number of diffusion steps $K$ & $200$ \\
Path length in training $T$ & $32$ \\
Classifier-free guidance scale $w$ & $1.2$ \\
End effector loss weight $w_{ee}$ & $10^{-3}$ \\
Collision loss weight $w_c$ & $10^{-3}$ \\
Neural network architecture & Convolutional U-Net \\
Number of model parameters & $\sim$$60\textrm{M}$ \\
Optimizer & Adam \\
Learning rate & $2\cdot10^{-4}$ \\
Batch size & $32$ \\
Number of training steps & $200\textrm{k}$ \\ \bottomrule
\end{tabular}
\end{table}

Our code is entirely accessible at the following location: \texttt{https://github.com/MSNikken/diffusion}

\subsection{Training Data}
In this work, we aim to create a path planner that is not robot-specific, nor requires expert examples, while still being able to produce optimal paths. To showcase the generalization capabilities of the DDP, the training dataset is synthetic and consists only out of straight lines/paths with associated returns. To create a path in the training dataset, two random poses are sampled uniformly from a 3D workspace. The remaining poses of each path are obtained by interpolating the positions and orientations between the two extreme of the trajectory. Note that these paths can be anywhere in the workspace, even in collision with obstacles. The dataset contains 60,000 paths.

In order to train the network to be able to avoid obstacles, the paths are associated with a return value \eqref{eq:returns}. To compute the return, we consider two different reward functions, namely a \textit{dense} and a \textit{sparse} reward function. The dense reward penalizes not only collisions with obstacles, but also the distance from the nearest obstacle.  Using the position of the nearest point of the nearest obstacle $\bm{p}_\text{obst}$, the dense reward can be written then:
\begin{equation}\label{eq:r_dense}
    r_{t}^{\text{dense}}\left(H_t\right) =
    \begin{cases}
        -1 & \text{when in collision}\\
        -e^{-a\cdot \lVert \bm{p}_\text{obst} - \bm{p}_t\rVert_2} & \text{otherwise,}
    \end{cases}
\end{equation}
where $a=46$ is a scaling parameter, $\bm{p}_t$ is the current end-effector position, and $\lVert \bm{p}_\text{obst} - \bm{p}_t \rVert_2$ is the Euclidean norm of $\bm{p}_\text{obst} - \bm{p}_t$. 
The sparse reward instead only penalizes collisions: 
\begin{equation}\label{eq:r_sparse}
    r_{t}^{\text{sparse}}\left(H_t\right) =
    \begin{cases}
        -1 & \text{when in collision,}\\
        0 & \text{otherwise.}
    \end{cases}
\end{equation}
In the following experiments, we consider static obstacles, a discount factor $\gamma=0.99$.

\subsection{Real-world Experiments}\label{subsec:realworld_exp}
To validate the DDP, different path planning tasks are performed with a Franka Emika Panda robot. Two different environments are created for validating the planner. For both environments, we define a cuboidal region in which some obstacles are placed. All training data is generated within this region. The first environment present a single cuboid obstacle (see Figure \ref{fig:real_world1}). The second environment contains two cuboid obstacles; a big cuboid and a small cube placed on top of it. The starting and goal poses are chosen such that no collision-free straight path exists between the start and the goal.

We test different strategies of conditional sampling. Firstly, we perform experiments using classifier-free guidance, using planners that were trained with dense rewards and planners trained with sparse rewards. Secondly, in order to make comparisons, we use the cost-guided conditional sampling approach presented in \cite{Carvalho2023MotionModels}. The cost function used in the experiments is the sum of two relevant cost terms used in \cite{Carvalho2023MotionModels}, namely the end-effector cost and the collision cost. The end-effector cost is given by:
\begin{equation}
    g_{ee}\left(H_t\right) = d_{SE(3)}\left(H_t, H_g\right)\, ,
\label{eq:ee_cost}
\end{equation}
where \[ d_{SE(3)}\left(H_t, H_g\right) = \lVert \bm{p}_t - \bm{p}_g\rVert^2_2 + \lVert LogMap\left(\bm{R}^T_t\bm{R}_g\right)\rVert^2_2.\] The $LogMap(\cdot)$ is a function from Lie Theory \cite{Sola2018ARobotics} that can be used to quantify the difference between two orientations.
The collision cost is given by:
\begin{equation}
    g_{c}\left(H_t\right) =
    \begin{cases}
        -d(\bm{p}_t) & \text{if } d(\bm{p}_t) \leq 0 \\
        0 & \text{if } d(\bm{p}_t) > 0
    \end{cases}
    \, ,
\end{equation}
where $d\left(\bm{p}_t\right)$ is a differentiable signed-distance function from the position $\bm{p}_t$ to the surface of the nearest obstacle, defined to be positive when $\bm{p}_t$ is outside the obstacle.
The total cost function is:
\begin{equation}
    J\left(\bm{\tau}\right) = \sum^{T-1}_{t=1} w_{ee} g_{ee}\left(H_t\right) + w_c g_c\left(H_t\right)\, ,
\label{eq:cost_guidance}
\end{equation}
with the weighting factors $w_{ee}$ and $w_{c}$.
Thirdly, we experiment with a novel way of sampling through combining classifier-free sampling with cost-guided sampling. In total, this results in five combinations for conditional sampling:

\begin{enumerate}[label=(\roman*)]
    \item \textit{Classifier-free guidance with dense rewards};
    \item \textit{Classifier-free guidance with sparse rewards};
    \item \textit{Cost guidance} using \eqref{eq:cost_guidance};
    \item \textit{Classifier-free guidance with dense rewards and cost guidance} (using only the end-effector cost \eqref{eq:ee_cost});
    \item \textit{Classifier-free guidance with sparse rewards and cost guidance} (using only the end-effector cost \eqref{eq:ee_cost}).
\end{enumerate}


The path planner is deployed in a receding-horizon configuration as explained in Section \ref{sec:closed-loop}. The following hyperparameters are chosen: the planning horizon $h=128$, number of steps tracked per execution $m=64$, number of paths produced per batch $n=5$, returns condition with dense rewards $c_{\textrm{dense}}=-0.01$, returns condition with sparse rewards $c_{\textrm{sparse}}=0$. In inpainting, the goal state is repeated five times at the end of the path (see Section \ref{subsec:repeated_inpainting}). Of the five paths generated per batch, the path with the lowest returns calculated with dense rewards is selected for tracking, regardless of the conditional-sampling method that is used. The generated paths are tracked using a Cartesian impedance control. We keep the the orientation of the end effector fixed to prevent the robot arm from twisting itself into a configuration that cannot be tracked by the low-level impedence controller anymore due to joint limits.

\section{Results and Discussions}
In this section, we show the results achieved with DDP. In particular, we study the impact of the choice of horizon and returns when conditionally-generating paths (Section \ref{subsec:impact_horizon_return}), the effect of repeatedly inpainting the goal state (Section \ref{subsec:repeated_inpainting}), and the performance of the DDP in real-world robotics tasks with different conditional-sampling approaches (Section \ref{subsec:realworld_results}).

\begin{figure*}[h!]
     \centering
     \begin{subfigure}[b]{0.49\textwidth}
         \centering
         \includegraphics[width=\textwidth]{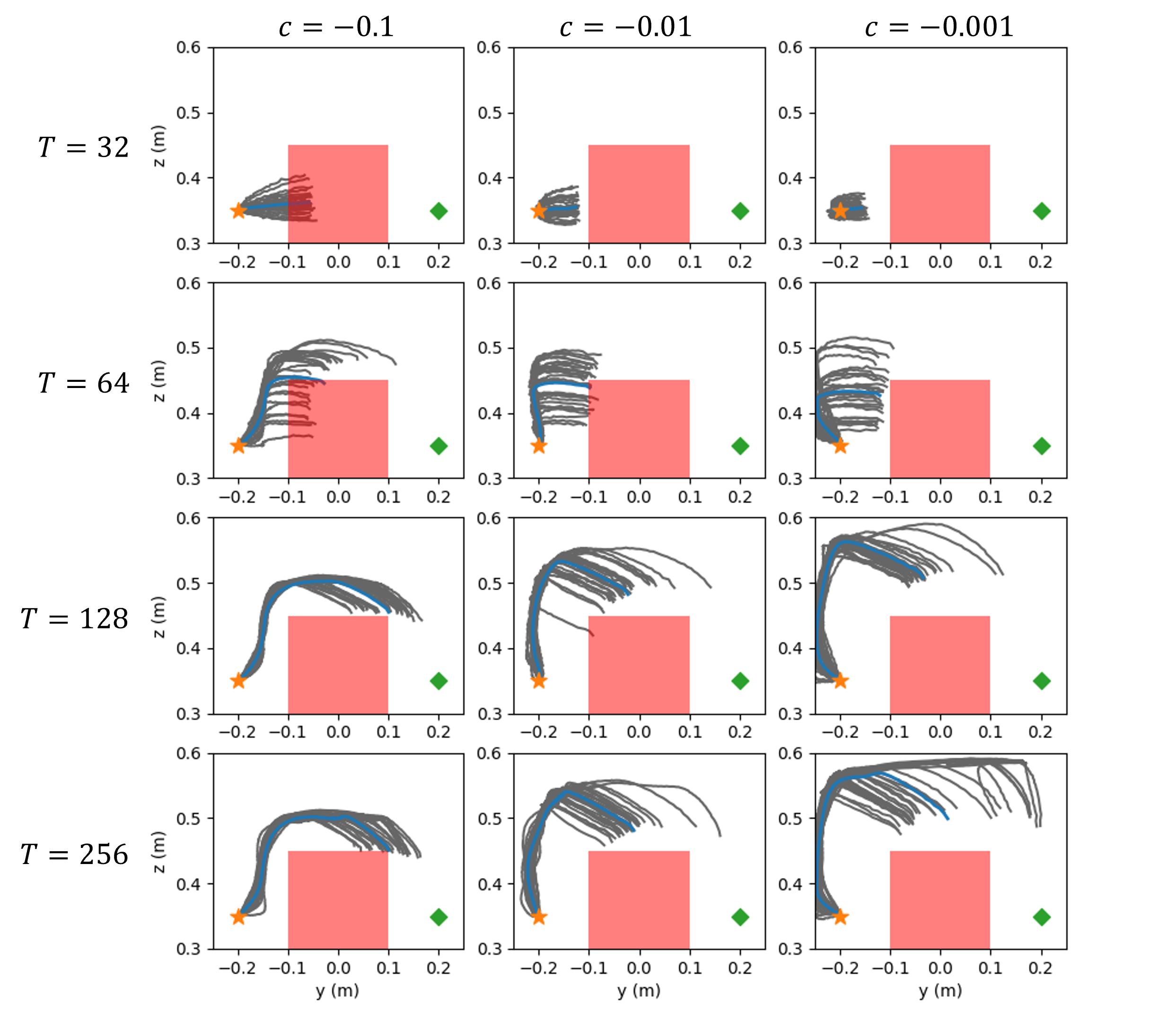}
         \caption{}
         \label{fig:par_dense}
     \end{subfigure}
     \begin{subfigure}[b]{0.49\textwidth}
         \centering
         \includegraphics[width=0.275\textwidth]{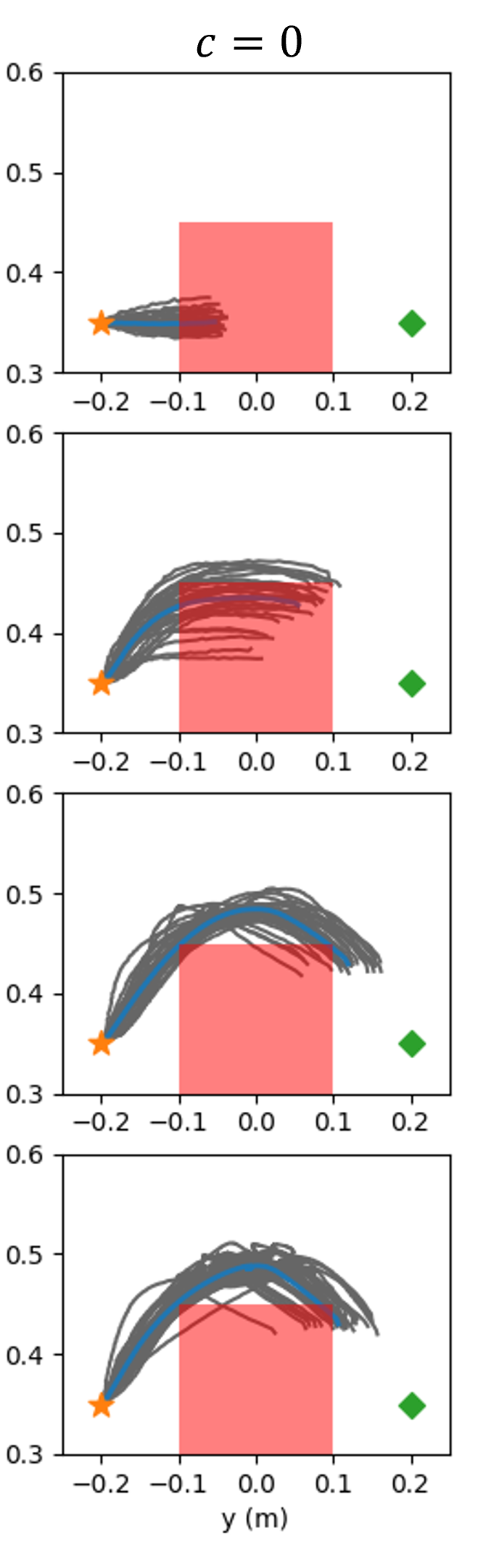}
         \caption{}
         \label{fig:par_sparse}
     \end{subfigure}
        \caption{Open-loop path generation with classifier-free guidance from the star to the diamond for various horizons $T$ and conditioned on various returns $c$. In (a), dense rewards are used and the returns are varied. In (b) sparse rewards are used and the returns are set to $c=0$. We generate thirty trajectories per combination of horizon and return and we highlight the mean trajectory with a blue line.}
        \label{fig:par_sweep}
\end{figure*}

\subsection{Impact of Horizon and Returns}\label{subsec:impact_horizon_return}
Since the training dataset contains paths in the entire workspace, even inside obstacles, the avoidance of obstacles is entirely achieved through conditional sampling. When using classifier-free guidance, these conditions are learned from the training data, which only consists out of straight lines. The extent to which the planner can generate paths that avoid an obstacle in between the start pose and goal pose depends on the DDPM's generalization capabilities, as well as on the hyperparameters used in inference. In particular, we focus our studies on two important hyperparameters: (i) the planning horizon and (ii) the (expected) return on which the sample is conditioned. We consider a simple scenario in which the DDP is used in open-loop configuration to inpaint a path between a starting pose and a goal pose, while an obstacle is positioned in between. Figure \ref{fig:par_sweep} shows simulations of the paths for various combinations of horizon and return values. For the model trained on sparse rewards, only a return value of zero is considered, since this is the only value that represents a collision-free path. Therefore, we effectively only sweep over the horizon in the sparse-reward case.
From Figure \ref{fig:par_sweep}, it is evident that in all configurations, coherent and nontrivial paths are generated. These paths start at the intended starting pose and tend towards the goal pose, even for horizons that are much longer than seen in training (see Table \ref{tab:settings}). The paths that were generated using sparse rewards are close to the obstacles (Figure \ref{fig:par_sparse}), whereas for dense rewards, the distance to the obstacle is strongly influenced by the rewards (Figure \ref{fig:par_dense}). 

For the model trained with dense rewards, collisions are avoided when conditioning on high rewards, but choosing them too high can lead to overly conservative paths. However, with sparse rewards, there is no control over the trade-off between generating a path that closely approaches the goal state and steering clear of the obstacle, which leads to many paths slightly violating the condition of obstacle avoidance, especially around the corners of the cuboid. For both types of rewards, the model is able to generate paths that are not in collision with the obstacle, when using a sufficiently long horizon ($\geq 64$). However, there is consistent gap is always present between the final planned state and the goal state. This issue is not merely due to the horizon being too short, and even doubling the horizon from $128$ to $256$ hardly improves this connection.

\subsection{Inpainting with repeated goal states}\label{subsec:repeated_inpainting}
\begin{figure}[h!]
    \centering
    \begin{subfigure}[b]{0.18\textwidth}
         \centering
         \includegraphics[width=\textwidth]{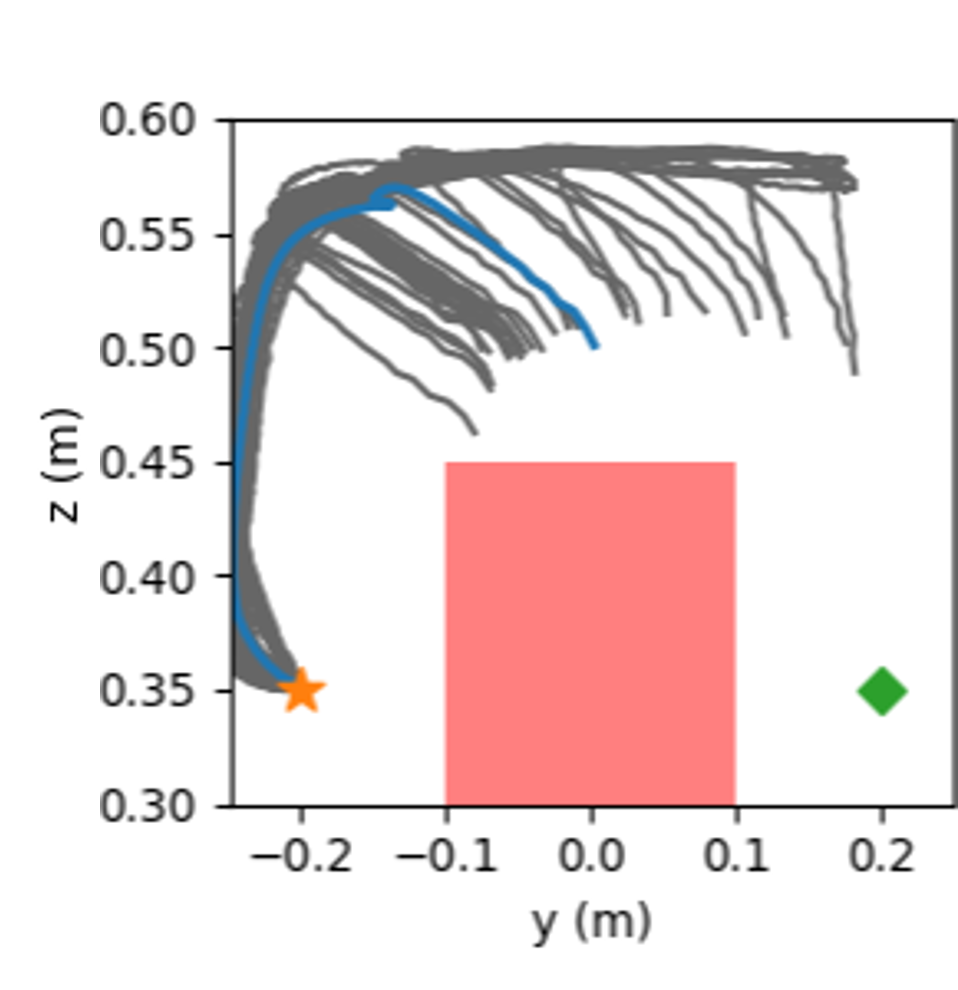}
         \caption{}
         \label{fig:inpaint_rep1}
    \end{subfigure}
    \begin{subfigure}[b]{0.18\textwidth}
         \centering
         \includegraphics[width=\textwidth]{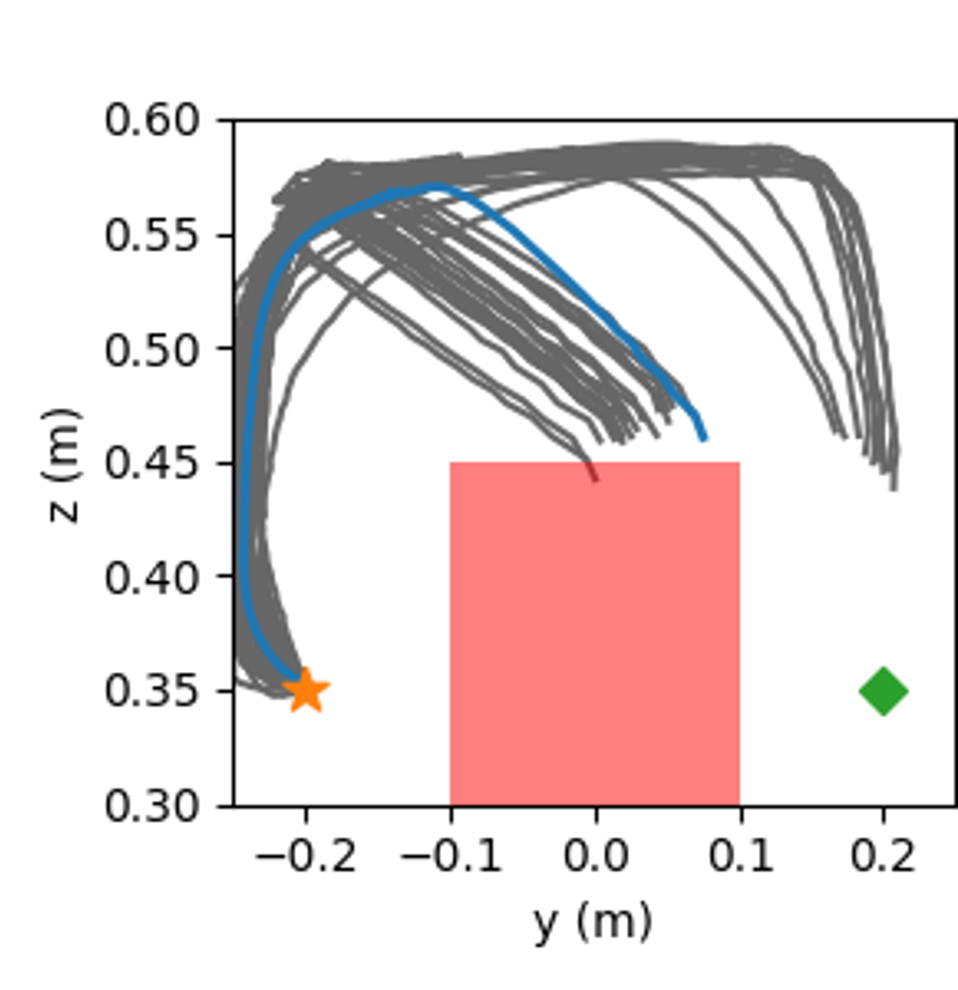}
         \caption{}
         \label{fig:inpaint_rep2}
    \end{subfigure}
    \begin{subfigure}[b]{0.18\textwidth}
         \centering
         \includegraphics[width=\textwidth]{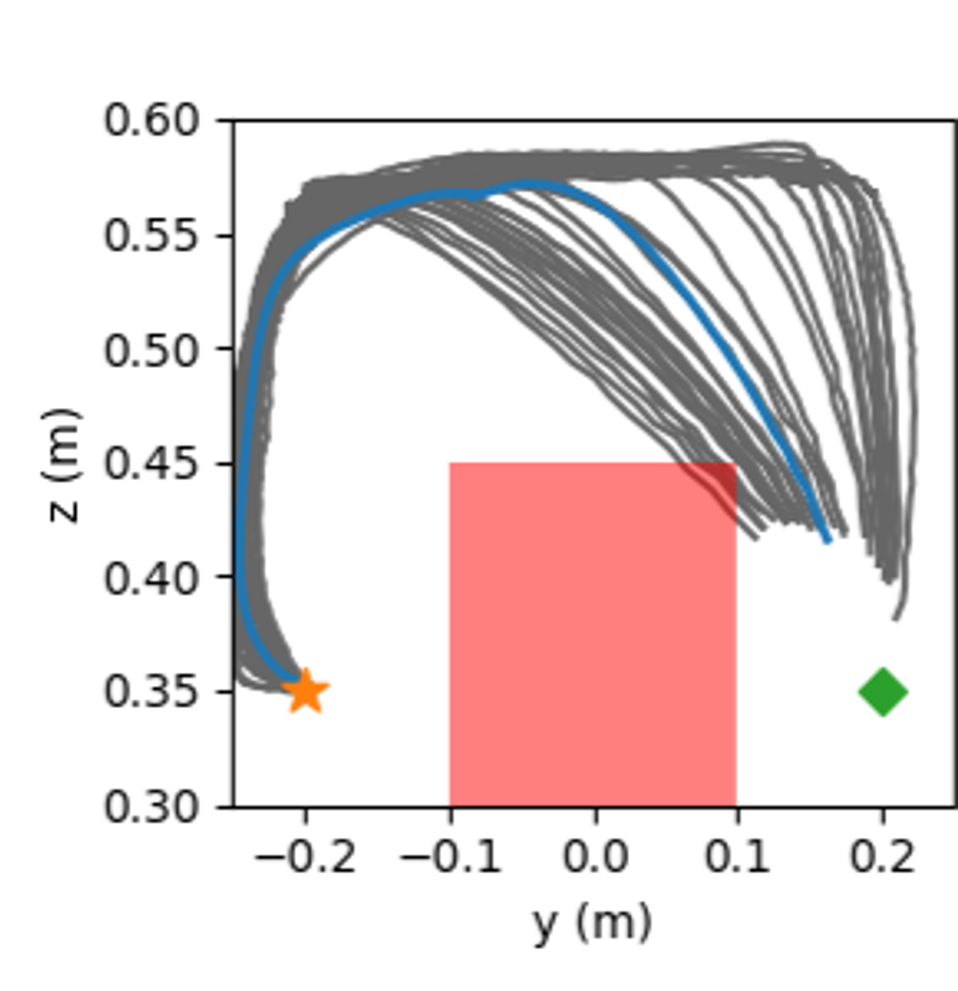}
         \caption{}
         \label{fig:inpaint_rep5}
    \end{subfigure}
    \begin{subfigure}[b]{0.18\textwidth}
         \centering
         \includegraphics[width=\textwidth]{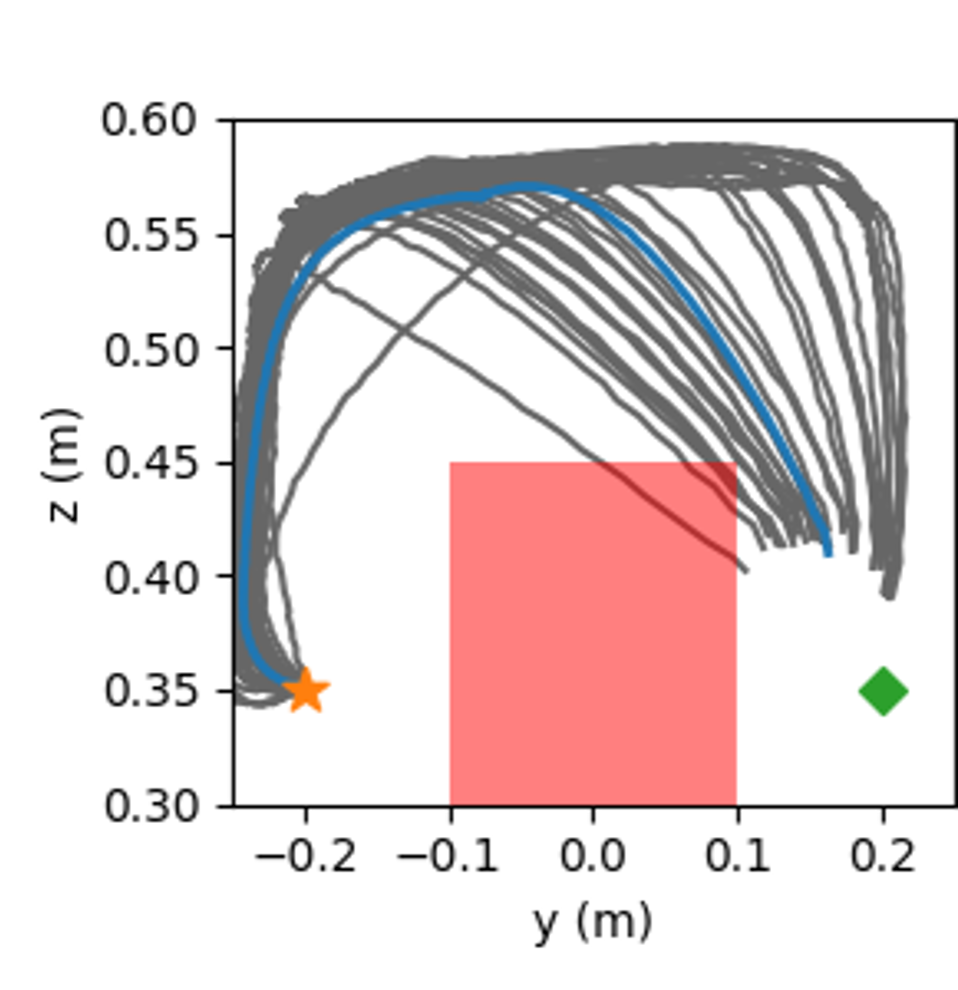}
         \caption{}
         \label{fig:inpaint_rep10}
    \end{subfigure}
    \caption{Open-loop path generation with classifier-free guidance with a varying number of repetitions of goal state inpainting. The number of repetitions are $1,2,5 \text{, and }10$, for (a), (b), (c), and (d), respectively. We generate thirty trajectories per combination of horizon and return and we highlight the mean trajectory with a blue line.}
    \label{fig:multiple_inpainting}
\end{figure}

A shown in Figure \ref{fig:par_sweep}, the generated paths generally do not typically connect well to the goal pose. We hypothesize that long horizons make the single goal state have relatively little weight in the reverse process, causing the gap at the end of the path. To improve upon this issue, insted of setting the last state of the trajectory equal to the goal state, we set the last $i$ states to be the goal state. We test this approach using the DDP trained with dense rewards,  horizon length of $T=256$ and conditioned on a return value of $c=-0.001$, which are the same parameters as used in the bottom-right of Figure \ref{fig:par_dense}. Simulations of the resulting paths are shown in Figure \ref{fig:multiple_inpainting}.

Figure \ref{fig:multiple_inpainting} shows that repeating the goal state in inpainting may be a useful trick. Even adding a single additional goal state to the reverse process already made the paths approach the goal more closely. However, there appears to be a limit, as inpainting with ten (or more) copies of the goal state yields no further improvement compared to using five states.

\subsection{Real-world Results}\label{subsec:realworld_results}

\begin{figure}[h!]
    \centering
    \includegraphics[width=1.0\linewidth]{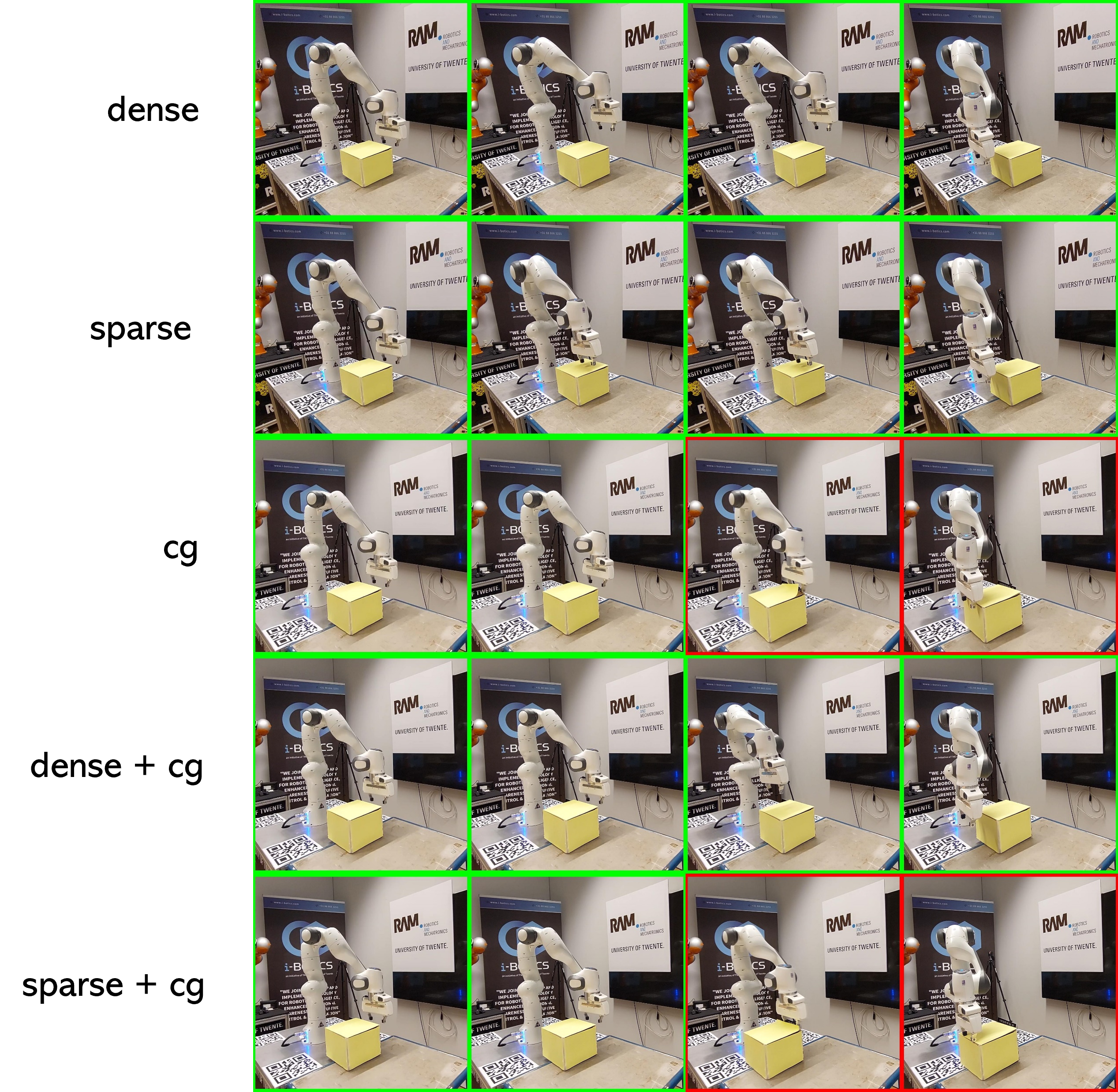}
    \caption{
        Frames from recordings of using various conditional sampling techniques (see Section \ref{subsec:realworld_exp}) to avoid a single obstacle. Green edges indicate that the robot is not in collision, whereas red edges indicate that the robot is in collision.
    }
    \label{fig:real_world1}
\end{figure}

We performed experiments in two different scenarios: the first with a single obstacle (for which Figure \ref{fig:real_world1} shows some frames); and the second with two obstacles, which forces the robot to generate more complicated 3D paths to avoid collisions. Videos of our experiments can be found \href{https://1drv.ms/f/c/4b7f21802058fb05/Eq4SFEAvfIlIhsqF6RoGT1ABgGEJd0UXHPpkpZlwmFt2mA?e=AVZqCj}{here}. We compare five different conditional-sampling strategies as described in Section \ref{subsec:realworld_exp}. 
The performed experiments confirmed that the DDP, trained only on low-quality synthetic data, is capable of repeatedly generate collision-free path in both test scenarios with different state and goal poses. The DDP demonstrated to be an extremely flexible framework that can exploit different configurations and conditional sampling strategies, as discussed in the previous section. 


We list some observations that we consider significant and potentially useful for practitioners.

With reference to Figure \ref{fig:real_world1}, the DDP trained with sparse rewards (sparse) resulted in paths much closer to the obstacles compared to the DDP trained dense rewards (dense), regardless of the inclusion of cost guidance (cg). However, in the second scenario with two obstacles, using only classifier-free guidance with dense rewards often results in lingering in place before crossing from one side of the smaller obstacle to the other. It is worth mentioning that our reward functions, i.e., sparse and dense, only penalizes collisions or distance to obstacles, respectively, and reaching the goal state is controlled by the fixing the value of the final state(s) of the generated trajectory (see Section \ref{subsec:repeated_inpainting}). Thus, except when utilizing cost guidance, the DDP is not explicitly conditioned to generate the shortest collision-free path from the initial to the goal state, but rather to generate an ``acceptable", collision-free path.

We observed that the DDP with classifier-free guidance with dense rewards can be improved by using the cost function guidance (dense + cg). In particular, in the scenario with two obstacles, the DDP without cost function guidance would often linger in place to avoid moving closer to the obstacles. Adding the cost guidance allows for solving this shortcoming by pushing the poses to be close to the goal pose.

However, the cost guidance alone (cg) and the classifier-free DDP with sparse rewards combined with cost guidance (sparse + cg) does not always allows for the generation of collision-free paths. These two approaches occasionally fail because there is no way promote to keep a minimum safe distance to the obstacles, beyond the edge of the obstacle, which opens the possibility of the end-effector cost bringing the path into collision.  Additionally, the cost guidance seems to be very sensitive to the choice of coefficients $w_{ee}$ and $w_c$ and careful tuning is required to achieve good results.

\section{Conclusion and Future Work}
In this work, we Denoising Diffusion Planner (DDP), an approach to design a path planner for a robotic end effector using DDPMs that does not use expert demonstrations, but only low-quality and synthetic data. Trained only with straight lines, the DDP can produce complex paths that avoid obstacles by utilizing the generalization capabilities of the DDPM and different conditional-sampling strategies. Additionally, deploying the DDP in a closed-loop receding-horizon control scheme and continuously update the plan allows for reaching difficult targets that cannot be reached in an open-loop configuration.

Future research will further investigate the trade-off between return conditioning, goal state inpainting, and planning horizon, as our results revealed these three elements crucial for generating high-quality paths. Additionally, it is essential to investigate strategies to reduce the need for (high-quality) data by exploiting physical priors \cite{Shu2023AReconstruction,Bastek2024Physics-InformedModels} or known geometries and symmetries  \cite{Yim2023SE3Generation, Jiang2023SEEstimation}.

\bibliographystyle{unsrt}  
\bibliography{references} 

\end{document}